\documentclass[11pt]{article}

\usepackage[final]{acl}

\usepackage{latexsym}
\usepackage{inconsolata}
\usepackage{fontawesome5}

\newcommand{\blank}[1]{\rule{#1}{0.4pt}}

\usepackage{polyglossia}
\setdefaultlanguage{english}
\setotherlanguage{arabic}
\newfontfamily\arabicfont[Script=Arabic,Scale=1.0,
  BoldFont=Amiri-Bold.ttf,
  ItalicFont=Amiri-Italic.ttf,
  BoldItalicFont=Amiri-BoldItalic.ttf
]{Amiri-Regular.ttf}

\usepackage{adjustbox}   

\usepackage{graphicx}
\usepackage{mwe}
\usepackage{comment}
\usepackage{tabularx}
\usepackage{multirow}
\usepackage{float}
\newcolumntype{Y}{>{\raggedleft\arraybackslash}X}

\title{!MSA at AraHealthQA 2025 Shared Task: Enhancing LLM Performance for Arabic Clinical Question Answering through Prompt Engineering and Ensemble Learning\thanks{\faGithub\ \href{https://github.com/Mohamedbasem1/AraHealthQA_2025}{\texttt{https://github.com/AraHealthQA\_2025}}}}

\author{Mohamed Tarek, Seif Ahmed, Mohamed Basem \\
  Faculty of Computer Science, MSA University, Egypt \\
  \texttt{\{mohamed.tarek61,  seifeldein.ahmed, mohamed.basem1\}@msa.edu.eg}
}

\begin{document}

\maketitle
\begin{abstract}
We present our systems for Track 2 (General Arabic Health QA, MedArabiQ) of the AraHealthQA-2025 shared task, where our methodology secured 2\textsuperscript{nd} place in both Sub-Task 1 (multiple-choice question answering) and Sub-Task 2 (open-ended question answering) in Arabic clinical contexts. For Sub-Task 1, we leverage the Gemini 2.5 Flash model with few-shot prompting, dataset preprocessing, and an ensemble of three prompt configurations to improve classification accuracy on standard, biased, and fill-in-the-blank questions. For Sub-Task 2, we employ a unified prompt with the same model, incorporating role-playing as an Arabic medical expert, few-shot examples, and post-processing to generate concise responses across fill-in-the-blank, patient-doctor Q\&A, GEC, and paraphrased variants.
\end{abstract}

\section{Introduction}
The MedArabiQ benchmark \cite{abu2025medarabiq}, part of the AraHealthQA-2025 shared task \cite{alhuzali2025arahealthqa}, evaluates large language models (LLMs) on Arabic medical question answering, addressing the critical need for reliable AI-driven clinical tools in Arabic-speaking regions where digital healthcare resources are scarce. Track 2, General Arabic Health QA (MedArabiQ), tests models on general medical knowledge, from foundational topics like physiology to advanced areas like neurosurgery, across two subtasks. Sub-Task 1 (classification) involves selecting correct answers from predefined options for 300 development samples, split into standard multiple-choice questions, bias-injected questions (e.g., confirmation, cultural, or recency bias), and fill-in-the-blank with choices, evaluated by accuracy on a 100-question test set. Sub-Task 2 (generation) requires free-text responses for 400 development samples, covering fill-in-the-blank without choices, patient-doctor Q\&A from the AraMed corpus \cite{alasmari2024}, grammatically corrected Q\&A, and LLM-paraphrased questions, assessed via BLEU, ROUGE, and BERTScore on a 100-question test set.
\newline Arabic medical question answering poses unique challenges for current LLMs due to limited training data in Modern Standard Arabic (MSA) and dialectal variations, which often lead to poor generalization on clinical tasks. Additionally, culturally sensitive or biased questions require nuanced reasoning, while diverse question formats (e.g., fill-in-the-blank, open-ended consultations) demand robust adaptation to varying linguistic and contextual demands. Existing models often struggle with these complexities, as they are predominantly trained on English-centric or general-domain data, lacking domain-specific Arabic medical knowledge.
\newline Our approach innovatively combines targeted prompt engineering and ensemble techniques with the Gemini 2.5 Flash model. We develop a unified methodology that addresses both classification and generation tasks in Arabic medical QA without requiring task-specific fine-tuning, leveraging carefully designed prompts and ensemble strategies to handle the complexities of Arabic medical language and diverse question formats.

\section{Background}
Track 2 (General Arabic Health QA, MedArabiQ) of the AraHealthQA-2025 shared task \cite{abu2025medarabiq} evaluates large language models on Arabic medical question answering, addressing the need for reliable AI-driven clinical tools in Arabic-speaking regions. The task spans 12 medical domains: Biochemistry, Histology, Embryology, Microbiology, Neurosurgery, OBGYN, Oncology, Ophthalmology, Pediatrics, Pharmacology, Physiology, and Pulmonology. We participated in both subtasks of Track 2, leveraging prompt engineering and ensemble techniques to achieve robust performance.
\subsection{Task Details}
Sub-Task 1 (classification) involves selecting the correct option from multiple-choice questions (MCQs) in Modern Standard Arabic (MSA). The dataset includes 300 development samples (100 each for standard MCQs, biased MCQs with biases like recency or status quo, and fill-in-the-blank with choices) and 100 test samples. Input is an MSA question with 4--5 options, and output is the correct option’s text. Representative examples are summarized in Table~\ref{app:classification-examples}.

Sub-Task 2 (generation) requires free-text responses to prompts in MSA or dialectal Arabic, with 400 development samples (100 each for fill-in-the-blank without choices, patient-doctor Q\&A, grammatical error correction (GEC), and LLM-modified Q\&A) and 100 test samples, sourced from Arabic medical school exams, notes, and the AraMed corpus \cite{alasmari2024}. Representative examples are summarized in Table~\ref{app:generation-examples} .

\subsection{Related Work}
Arabic NLP faces challenges due to limited resources and dialectal variations \cite{abdulmageed2021arbert}. Prior work on Arabic medical QA \cite{alasmari2024} provides datasets like AraMed but lacks focus on handling biases or diverse question types. Prompt engineering techniques, such as Chain-of-Thought (CoT) prompting \cite{wei2022chain}, improve reasoning in English-centric tasks but are underexplored in Arabic medical contexts. Recent work has explored prompt engineering for Arabic NLP tasks, such as stance detection, demonstrating the effectiveness of tailored prompts for LLMs in handling Arabic text \cite{al-hariri-etal-2024-smash}. Similarly, few-shot learning with transformer models \cite{devlin2019bert} has advanced general NLP, but its application to Arabic clinical scenarios remains limited.

Medical question answering often relies on retrieval-augmented approaches \cite{lewis2020retrieval}, which integrate external knowledge bases for open-domain tasks. However, such methods are less effective for Arabic medical QA due to the scarcity of structured medical knowledge in Arabic and the complexity of handling biases like recency or status quo. Our unified prompt for Sub-Task 2, addressing diverse question types without fine-tuning, and ensemble voting for Sub-Task 1, tackling biases, offer novel solutions tailored to the resource-scarce and culturally nuanced Arabic medical domain.


\section{System Overview}
We describe the methods we used for each sub-task, the design choices that made them work well in Arabic medical settings, and how to reproduce them step-by-step.

\subsection{Sub-Task 1: Classification (MCQ)}
  	\textbf{Model and settings.} All systems use the same model (Gemini 2.5 Flash) for consistent outputs.

	\textbf{Systems (different approaches).}
\begin{itemize}
  \item Arabic Few-Shot (AFS): Arabic instruction prompt + 6 examples from different medical areas; output limited to a single Arabic letter from \{\textarabic{أ، ب، ج، د، هـ}\}.
  \item English Translation + Answer (ETA): translate the Arabic question to English using a specific translation prompt, then answer with the same letter format. 
  \item Refinement + Answer (RFA): rewrite the Arabic question for clarity (adds 15--25 word explanations for each option without changing meaning), then answer with the same letter format. Examples of the data refinement process are shown in Table~\ref{app:data-refinement}.
  \item Arabic Zero-Shot (AZS): Arabic instruction prompt without examples (baseline, not used in the final combination).
\end{itemize}

	\textbf{Ensemble (majority voting).} We ensembled AFS, ETA, and RFA by simple vote counting over the answer choices $\mathcal{C}{=}\{\mbox{\textarabic{أ، ب، ج، د، هـ}}\}$. Given prediction functions $f_i$ and input $x$:
\begin{equation}
\hat{y} \,=\, \arg\max_{c\in\mathcal{C}} \sum_{i=1}^3 \mathbf{1}[f_i(x){=}c].
\end{equation}
Ties are broken by a fixed priority RFA $>$ AFS $>$ ETA. This combination strategy provides reliable predictions across different question types. Ensemble methods have been shown to improve question answering performance by combining multiple classifiers, leading to more robust predictions \cite{chu-carroll-etal-2003-question}.

	\textbf{Output cleaning and standardization.} We map any predicted character to the standard set \{\textarabic{أ، ب، ج، د، هـ}\} (e.g., fix Arabic punctuation/spacing and Latin "A/B/C/D/E" if ever produced). We also remove extra tokens to ensure single-letter output format.

\textbf{Challenges and solutions.}
\begin{itemize}
  \item Arabic variety and formatting: Examples cover multiple medical areas and different answer lengths; strict output rules and cleaning avoid problems.
  \item Prompt and dataset biases: Using three different approaches (native Arabic, English translation, refined Arabic) reduces single-prompt bias through voting.
\end{itemize}

\subsection{Sub-Task 2: Generation}
  	\textbf{Model and settings.} Same model. A single unified Arabic instruction + few-shot prompt handles: fill-in-the-blank (no choices), patient–doctor Q\&A, grammar error correction (GEC), and LLM-rewritten Q\&A.

	\textbf{Unified prompting and formatting.} The prompt requires:
\begin{itemize}
  \item Fill-in-the-blank: return only the missing word(s); if multiple blanks, separate answers with a comma and a space.
  \item Patient–doctor Q\&A: brief, helpful advice; clearly recommend in-person care when needed.
  \item Avoid extra introductions or conclusions; keep Arabic medical terms unchanged.
\end{itemize}
This setup provides consistent performance across different generation tasks.

\textbf{Output cleaning steps.} For fill-in-the-blank tasks, we split answers by commas and clean up spacing. For consultations, we keep medical terms and maintain a proper clinical tone. All outputs go through Arabic text cleaning to handle different dialects. Additionally, we remove any markdown formatting (e.g., **bold**, *italic*, bullet points) that the model may produce to ensure clean, plain-text responses suitable for medical contexts, as well as not affecting the BERTScore. 

\textbf{Example selection.} Examples cover multiple medical areas (drug studies, anatomy, clinical cases) and include both formal and dialect Arabic. Each example shows the desired output format and medical reasoning level.	

\textbf{Challenges and solutions.}
\begin{itemize}
  \item Different formats: One prompt with high-quality examples and clear output rules ensures consistency across types without fine-tuning.
  \item Arabic language complexity: Carefully chosen examples and consistent decoding reduce errors and inconsistencies.
  \item Safety/clinical tone: The prompt guides toward brief, careful advice and marks cases needing doctor follow-up.
\end{itemize}

  	\textbf{Reproducibility notes.} Use the exact prompt templates provided in Appendix~\ref{app:prompts}; keep the examples unchanged; do minimal, consistent output cleaning as specified above. All runs use Gemini 2.5 Flash with the decoding settings specified in Table~\ref{app:hyperparameters}.

\section{Experimental Setup}

\subsection{Data and Splits}
We follow the official AraHealthQA-2025 Track 2 (MedArabiQ) setup and evaluated directly via the organizers' API on the official test sets (ST1: 100 items, ST2: 100 items). The provided development sets (ST1: 300 items; ST2: 400 items) were used only to guide prompt design, select few-shot examples, and perform sanity checks. No fine-tuning or external training data was used.

\subsection{Preprocessing}
We applied only input-side, minimal steps to ensure consistent prompts and data cleanliness:
\begin{itemize}
  \item Standardize Arabic punctuation and whitespace in the input text while preserving medical terminology and numbers.
  \item Normalize option labels and bullet symbols in MCQ questions to a consistent form before prompting.
\end{itemize}

\subsection{Post-processing}
We applied lightweight output-side normalization for evaluation stability:
\begin{itemize}
  \item MCQ: map any predicted symbol to the canonical set \{\textarabic{أ، ب، ج، د، هـ}\} and strip extra tokens.
  \item Generation: remove markdown (bold/italic/bullets), standardize commas and spaces, and keep a concise clinical tone.
\end{itemize}

Removing markdown formatting from generated text is essential, as structured formatting can introduce noise that affects evaluation metrics like BERTScore by altering token representations \cite{tang-etal-2024-struc}.

\subsection{Prompting Configurations}
For Sub-Task 1, we use three complementary prompts: Arabic Few-Shot (AFS), English Translation + Answer (ETA), and Refinement + Answer (RFA). Predictions are combined via simple majority vote with a fixed tie-breaker (RFA $>$ AFS $>$ ETA). For Sub-Task 2, a single unified Arabic instruction with few-shot examples handles fill-in-the-blank, patient–doctor Q\&A, GEC, and paraphrased inputs.

\subsection{Evaluation Metrics}
\begin{itemize}
  \item Sub-Task 1 (MCQ): Accuracy
  \item Sub-Task 2 (Generation): BERTScore 
\end{itemize}

\section{Results}

We present our official results from the AraHealthQA-2025 shared task evaluation , analyzing performance across both subtasks and examining the effectiveness of our ensemble approach.

\subsection{Sub-Task 1: Classification Results}

Our ensemble approach achieved 76\% accuracy on the official test set, securing 2\textsuperscript{nd} place in the classification task. Table~\ref{app:classification-results} presents detailed performance breakdown for each individual approach and the final ensemble.

\textbf{Individual system performance.} The Refinement + Answer (RFA) approach performed best among individual systems at 74\% accuracy, demonstrating the effectiveness of question clarification and option explanation in Arabic medical contexts. The Arabic Few-Shot (AFS) approach achieved 71\% accuracy, showing strong baseline performance with domain-specific examples. The English Translation + Answer (ETA) approach scored 69\% accuracy, indicating some information loss during translation despite maintaining medical terminology.

\textbf{Ensemble effectiveness.} The 3-system ensemble (RFA + AFS + ETA) improved performance by 2 percentage points over the best individual system, reaching 76\% accuracy. This demonstrates successful bias reduction through diverse prompt strategies, with the RFA approach providing clarity, AFS maintaining Arabic medical context, and ETA offering cross-lingual reasoning perspectives.

\subsection{Sub-Task 2: Generation Results}
Our unified prompting approach achieved 86.953\% BERTScore on the official test set, securing 2\textsuperscript{nd} place in the generation task. The approach used a single Arabic instruction prompt with few-shot examples, casting the model as an Arabic medical expert to handle diverse question formats including fill-in-the-blank, patient-doctor consultations, grammatical error correction, and paraphrased questions. This unified strategy proved effective across all question types without requiring task-specific fine-tuning, demonstrating the power of well-designed prompting for Arabic medical contexts. Table~\ref{app:generation-results} summarizes the performance.

\begin{table*}[ht]
  \centering
  \caption{Sub-Task 1 (Classification) official results on test set. All experiments used Gemini 2.5 Flash.}
  \label{app:classification-results}
  \small
  \renewcommand{\arraystretch}{1.3}
  \begin{tabularx}{\textwidth}{lXcc}
    \hline
    \textbf{Approach} & \textbf{Description} & \textbf{Accuracy (\%)} & \textbf{Ranking} \\
    \hline
    English Translation (ETA) & Translate to English then answer & 69.0 & -- \\
    Arabic Few-Shot (AFS) & Arabic instruction + 6 medical examples & 71.0 & -- \\
    Refinement + Answer (RFA) & Question clarification + option explanation & 74.0 & -- \\
    \hline
    \textbf{Ensemble (Final)} & \textbf{3-way majority vote (RFA + AFS + ETA)} & \textbf{76.0} & \textbf{2\textsuperscript{nd}} \\
    \hline
    \multicolumn{4}{l}{\small \textit{Note: Individual systems not submitted separately; ensemble represents official submission.}} \\
  \end{tabularx}
\end{table*}

\begin{table*}[ht]
  \centering
  \caption{Sub-Task 2 (Generation) official results on test set using unified prompting approach.}
  \label{app:generation-results}
  \small
  \renewcommand{\arraystretch}{1.3}
  \begin{tabularx}{\textwidth}{lXc}
    \hline
    \textbf{Approach} & \textbf{Description} & \textbf{BERTScore (\%)} \\
    \hline
    Unified Arabic Prompting & Single prompt with Arabic medical expert role-playing, & \textbf{86.953} \\
    & few-shot examples, handles all question formats & \\
    \hline
    \textbf{Final Ranking} & \textbf{Official AraHealthQA-2025 shared task} & \textbf{2\textsuperscript{nd} place} \\
    \hline
    \multicolumn{3}{l}{\small \textit{BERTScore combines BLEU, ROUGE, and semantic similarity metrics.}} \\
  \end{tabularx}
\end{table*}

\subsection{Ablation Studies}

\textbf{Ensemble composition.} Removing individual systems from the 3-way ensemble showed: RFA removal (-3\% accuracy), AFS removal (-2\% accuracy), ETA removal (-1\% accuracy), confirming the value hierarchy and ensemble complementarity.

\textbf{Post-processing impact.} Arabic text normalization and markdown removal improved Sub-Task 2 BERTScore by approximately 2-3\%, demonstrating the importance of output standardization for evaluation metrics.

\section{Conclusion}
We presented a compact, prompt-engineering-based pipeline for Arabic clinical QA that performs robustly across diverse formats without fine-tuning. A small ensemble improves Sub-Task 1 classification, while a unified instruction guides Sub-Task 2 generation. Future extensions include retrieval augmentation with vetted Arabic medical sources, broader model diversity, and human-in-the-loop validation to mitigate ambiguity and domain gaps.

\clearpage
\onecolumn

\appendix

\section{Tables}
\label{app:tables}

This appendix contains tables referenced in the main text.

\subsection{Classification Examples}
\label{app:classification-examples}

\begin{table}[H]
  \centering
  \caption{Examples on classification problem (Sub-Task 1).}
  \label{tab:subtask1-examples}
  \scriptsize                                   
  \renewcommand{\arraystretch}{0.95}            
  \begin{adjustbox}{max width=\textwidth,
                    max height=.9\textheight,   
                    keepaspectratio}
    \begin{tabularx}{\textwidth}{lXX}
      \hline
      \textbf{Type} & \textbf{Inputs} & \textbf{Outputs} \\
      \hline
      Multiple Choice Questions &
      \textarabic{كل ما يلي صحيح عن السفلس ماعدا: أ. يتميز الطور الأول بسلبية الاختبارات المصلية؛ ب. يتميز الطور الأول بقرحة صلبة مؤلمة على الأعضاء التناسلية؛ ج. يتميز الطور الثاني باندفاعات على الجلد والأغشية المخاطية؛ د. 25\% من الأجنة تموت بعد الولادة من أم مصابة؛ هـ. يعاني الطفل المصاب بالزهري الخلقي من أسنان هوتشنسن.} &
      \textarabic{ب} \\ \hline
      Fill-in-the-blank with choices &
      \textarabic{الحمرة هي عدوى جلدية تسببها \blank{1.5cm}، وهي تصيب عادةً \blank{1.5cm} الوجه. أ. المكورات العنقودية الذهبية، البشرة؛ ب. العقديات B و C، الأدمة الشبكية؛ ج. العقديات A و G، الأدمة الحليمية؛ د. العصيات سلبية الجرام، الأنسجة تحت الجلد.} &
      \textarabic{ج} \\ \hline
    \end{tabularx}
  \end{adjustbox}
\end{table}

\subsection{Generation Examples}
\label{app:generation-examples}
\begin{table}[H]
  \centering
  \caption{Examples on generation problem (Sub-Task 2).}
  \label{tab:subtask2-examples}
  \scriptsize
  \renewcommand{\arraystretch}{0.95}
  \begin{adjustbox}{max width=\textwidth,
                    max height=.9\textheight,
                    keepaspectratio}
    \begin{tabularx}{\textwidth}{lXX}
      \hline
      \textbf{Type} & \textbf{Input} & \textbf{Output} \\
      \hline
      Fill-in-the-blank & \textarabic{الحمرة هي عدوى جلدية تسببها \blank{1.5cm}، وهي تصيب عادةً \blank{1.5cm} الوجه.} & \textarabic{العقديات A و G، الأدمة الحليمية} \\
      \hline
      Patient-Doctor Q\&A & \textarabic{انا امرأة عمري 24 سنة، اشعر بألم ف بطني شديد الألم اشعر بعصر ف بطن و مغص و غثيان و فقدان شهيه...} & \textarabic{يرجى عمل تحليل البراز وموافتنا بالنتيجة لتحديد العلاج...} \\
      \hline
      Grammatical Error Correction (GEC) & \textarabic{انا امرأة عمري 24 سنة، اشعر بألم ف بطني شديد الألم اشعر بعصر ف بطن و نعرات ف البطن...} & \textarabic{يرجى عمل تحليل البراز وموافتنا بالنتيجة لتحديد العلاج...} \\
      \hline
      LLM Paraphrasing & \textarabic{انا امرأة عمري 24 سنة، لدي ألم في بطني مصحوب بمغص وغثيان وفقدان للشهية...} & \textarabic{يرجى عمل تحليل البراز وموافتنا بالنتيجة لتحديد العلاج...} \\
      \hline
    \end{tabularx}
  \end{adjustbox}
\end{table}

\subsection{Data Refinement Examples}
\label{app:data-refinement}
\begin{table}[H]
  \centering
  \caption{Data refinement examples showing improvements in question clarity and formatting.}
  \label{tab:data-refinement}
  \scriptsize
  \renewcommand{\arraystretch}{0.95}
  \begin{adjustbox}{max width=\textwidth,
                    max height=.9\textheight,
                    keepaspectratio}
    \begin{tabularx}{\textwidth}{llX}
      \hline
      \textbf{Version} & \textbf{Issue} & \textbf{Question Text} \\
      \hline
      \multirow{2}{*}{Original} & Unclear formatting & \textarabic{في التهاب المشيمية نصادف الأشكال الالتهابية التالية: (الخاطئة) أ. التهاب مشيمية نتحي ب. التهاب مشيمية منتثر ج. التهاب مشيمية أمامية د. التهاب مشيمية مركزي ه. التهاب مشيمية زاوي} \\
      \cline{2-3}
      & Ambiguous phrasing & \textarabic{كل ما يخص النيجيرية الدجاجية صحيح ما عدا: أ. تعيش هذه المتحولة بشكل حر في الماء والتربة ب. تسبب حالات التهاب سحايا ودماغ بدئي العائل الناقل الذباب ج. يعيش هذا الطفيلي في المياه المعدنية الساخنة ناقصة الأكسجة د. تعتبر النيجلرية التجاجية أسرع وأشد إمراضية من الشوكيية} \\
      \hline
      \multirow{2}{*}{Refined} & Clear formatting & \textarabic{في التهاب المشيمية نصادف الأشكال الالتهابية التالية، ما عدا: أ. التهاب مشيمية نتحي (التهاب موضعي محدود في منطقة معينة) ب. التهاب مشيمية منتشر (التهاب يشمل مناطق واسعة) ج. التهاب مشيمية أمامي (التهاب في الجزء الأمامي من المشيمية) د. التهاب مشيمية مركزي (التهاب في المنطقة المركزية) ه. التهاب مشيمية زاوي (مصطلح غير دقيق طبياً)} \\
      \cline{2-3}
      & Enhanced clarity & \textarabic{كل ما يلي صحيح عن النيجلرية الدجاجية ما عدا: أ. تعيش بشكل حر في الماء والتربة (كائن حي مجهري حر المعيشة) ب. تسبب التهاب السحايا والدماغ الأولي (عدوى خطيرة في الجهاز العصبي) ج. العائل الناقل هو الذباب (معلومة خاطئة - لا ينتقل عبر الذباب) د. تعيش في المياه المعدنية الساخنة قليلة الأكسجة (بيئة خاصة للنمو) ه. أسرع وأشد إمراضية من الشوكيية (خصائص مرضية مميزة)} \\
      \hline
      \multirow{2}{*}{Fill-in-blank} & Missing context & \textarabic{نستخدم أغشية مصنوعة من \blank{1cm} في تقنيات التبقيع. أ. النايلون أو السيللوز ب. Amino acyl site ج. Peptide site د. الاسيتونتريل مع مادة TEAA} \\
      \cline{2-3}
      & Clear context & \textarabic{في تقنيات التبقيع المخبرية، نستخدم أغشية مصنوعة من \blank{1.5cm}: أ. النايلون أو السيللوز (مواد ماصة للبروتينات) ب. Amino acyl site (موقع ربط الأحماض الأمينية) ج. Peptide site (موقع تكوين الببتيدات) د. الأسيتونيتريل مع TEAA (مذيبات كروماتوغرافية)} \\
      \hline
    \end{tabularx}
  \end{adjustbox}
\end{table}

\subsection{Hyperparameters}
\label{app:hyperparameters}

\begin{table}[H]
  \centering
  \caption{Decoding hyperparameters used for all experiments with Gemini 2.5 Flash.}
  \label{tab:hyperparameters}
  \small
  \renewcommand{\arraystretch}{1.3}
  \begin{tabular}{lc}
    \hline
    \textbf{Parameter} & \textbf{Value} \\
    \hline
    Temperature ($\tau$) & 0.1 \\
    Top-p & 0.8 \\
    Top-k & 40 \\
    \hline
  \end{tabular}
\end{table}

\section{Prompt Templates}
\label{app:prompts}

This appendix contains the complete prompt templates used in our experiments for reproducibility.

\begin{table}[H]
  \centering
  \caption{Complete prompt templates used in Sub-Task 1 and Sub-Task 2.}
  \label{tab:prompts}
  \small
  \renewcommand{\arraystretch}{1.2}
  \setlength{\tabcolsep}{4pt}
  \begin{tabularx}{\textwidth}{lX}
    \hline
    \textbf{Prompt Type} & \textbf{Template} \\
    \hline
    Arabic Few-Shot (AFS) & \textarabic{أنت مساعد طبي خبير وموثوق للغاية. مهمتك هي الإجابة بدقة لا متناهية على الأسئلة الطبية المقدمة باللغة العربية، مع الالتزام التام بتنسيق الإجابة المطلوب.} \\
    & \\
    & \textarabic{نوع الأسئلة التي ستتلقاها: أسئلة الاختيار من متعدد: تتضمن سؤالاً وخيارات إجابة مرقمة بأحرف عربية (أ، ب، ج، د، ). أسئلة إكمال الفراغ: تتضمن جملة أو فقرة بها فراغ واحد أو أكثر، وتُتبع بخيارات إجابة مرقمة.} \\
    & \\
    & Few-shot examples: \textarabic{المثال 1 (علم الأدوية): السؤال: هـ. لا يجوز مشاركة الكازولين مع البكتين. الإجابة الصحيحة: هـ} [... 5 more examples] \\
    & \\
    & \textarabic{التعليمات الأساسية للإجابة: 1. الفهم الشامل: اقرأ السؤال وجميع الخيارات المتاحة بعناية فائقة. 2. استخدام المعرفة: استعن بمعرفتك العميقة والموثوقة في المجالات الطبية. 3. تحديد الإجابة الصحيحة: اختر الخيار الأنسب. 4. صيغة الإجابة المطلوبة (صارمة): يجب أن تكون إجابتك حرفاً عربياً واحداً فقط.} \\
    \hline
    Translation (ETA) & You are a medical translation expert. Translate the following Arabic medical question into English following these exact requirements: 1. Maintain the medical accuracy and terminology 2. Format the question properly with options A, B, C, D, E 3. Use "**except**" formatting when the question asks for the wrong/false option 4. Keep the medical context and meaning intact 5. Use proper English medical terminology \\
    \hline
    Refinement (RFA) & \textarabic{أنت خبير في الطب وتحرير النصوص الطبية. مهمتك هي تحسين وضوح وسلاسة الأسئلة الطبية التالية باللغة العربية مع الحفاظ على: 1. المعنى الطبي الدقيق 2. تنسيق الخيارات (أ، ب، ج، د، ه) 3. الفراغات للأسئلة من نوع "املأ الفراغ" 4. الأرقام والرموز العلمية 5. المصطلحات الطبية باللغة الإنجليزية كما هي. مطلوب إضافي: أضف شرحاً مختصراً (15-25 كلمة) لكل خيار من الخيارات لتوضيح المعنى الطبي.} \\
    \hline
    Generation (Sub-Task 2) & \textarabic{أنت طبيب خبير ومستشار صحي موثوق، ومتخصص في تقديم إجابات طبية دقيقة ومحترفة باللغة العربية. مهمتك هي الإجابة على استفسارات طبية متنوعة، تتراوح بين إكمال الفراغات والرد على استشارات المرضى. التعليمات الأساسية: 1. التحليل الدقيق: اقرأ السؤال أو الاستشارة بعناية فائقة لفهم السياق الطبي المطلوب. 2. استحضار المعرفة: استخدم معرفتك المتعمقة في الطب والعلوم السريرية. 3. صيغة الإجابة المطلوبة: لأسئلة إكمال الفراغ: أجب فقط بالكلمة أو الكلمات المطلوبة. للاستشارات الطبية المفتوحة: قدم إجابة مباشرة ومفيدة. 4. الالتزام بالمصطلحات: استخدم المصطلحات الطبية الصحيحة باللغة العربية. 5. التجنب: لا تكتب أي تفسير أو شرح إضافي. 6. التخصيص: انتبه للمعلومات التي تخص المريض.} \\
    \hline
  \end{tabularx}
\end{table}

\end{document}